\documentclass{article}

    \PassOptionsToPackage{numbers, compress}{natbib}

\usepackage[preprint]{neurips_2020}
\usepackage[hyphens]{url}
\usepackage{hyperref}

\usepackage[utf8]{inputenc} 
\usepackage[T1]{fontenc}    
\usepackage{hyperref}       
\usepackage{url}            
\usepackage{booktabs}       
\usepackage{amsfonts}       
\usepackage{nicefrac}       
\usepackage{microtype}      
\usepackage{graphicx}
\usepackage{amsmath}
\usepackage{authblk}

\title{3DMaterialGAN: Learning 3D Shape Representation from Latent Space for Materials Science Applications}

 \author[1]{\textbf {Devendra K. Jangid}}
 \author[3]{\textbf {Neal R. Brodnik}}
 \author[1]{\textbf {Amil Khan}}
 \author[2]{\textbf {McLean P. Echlin}}
 \author[2]{ \textbf {Tresa M. Pollock }}
 \author[3]{\textbf {Sam Daly}}
 \author[1]{\textbf {\hspace{0.9cm}B. S. Manjunath }}

 \affil[1]{\footnotesize Electrical and Computer Engineering}
 \affil[2]{\footnotesize Materials Department}
 \affil[3]{\footnotesize Mechanical Engineering}
 \affil[ ]{\footnotesize University of California, Santa Barbara}
 \affil[ ]{\texttt {\{tresap, samdaly, manj\}@ucsb.edu}}

\begin{document}

\maketitle

\begin{abstract}
  In the field of computer vision, unsupervised learning for 2D object generation has advanced rapidly in the past few years.  However, 3D object generation has not garnered the same attention or success as its predecessor.  To facilitate novel progress at the intersection of computer vision and materials science, we propose a 3DMaterialGAN network that is capable of recognizing and synthesizing individual grains whose morphology conforms to a given 3D polycrystalline material microstructure.  This Generative Adversarial Network (GAN) architecture yields complex 3D objects from probabilistic latent space vectors with no additional information from 2D rendered images.  We show that this method performs comparably or better than state-of-the-art on benchmark annotated 3D datasets, while also being able to distinguish and generate objects that are not easily annotated, such as grain morphologies. The value of our algorithm is demonstrated with analysis on experimental real-world data, namely generating 3D grain structures found in a commercially relevant wrought titanium alloy, which were validated through statistical shape comparison.  This framework lays the foundation for the recognition and synthesis of polycrystalline material microstructures, which are used in additive manufacturing, aerospace, and structural design applications.

\end{abstract}

\section{Introduction}


The ability to generate 3D objects is of interest across many different fields because of its impact on a diverse application space. Much of this stems from the potential to synthesize realizable data that can be used to solve the small dataset problem where data is expensive, or to explore the unique complexities of a dataset beyond the sum of its parts. For example, there are countless possible chair designs, but we never look at a chair and go through a checklist of designs to verify that it is one. We simply let shape dictate function. In this same respect, a generative model should be able to learn the shape of an object and its constituent components within a particular context.

\textbf{Generative Models: } As more comprehensive datasets like ShapeNet from \citet{chang2015shapenet} and \citet{Wu2015} become available to the community, this will enable us to build better generative models to tackle the difficult problem of 3D object generation. Although there have been previous successful attempts that provide solutions to this task, the quality of the objects generated still require improvement, and many opportunities remain for the application of these models to real-world problems.


In this paper, we develop a novel generative architecture that provides promising results for learning the shape of volumetric data. Inspired by \citet{Goodfellow_NIPS2014_5423}, Generative Adversarial Networks (GANs) try to generate new data based on a training data distribution. Building on this foundational idea and applying concepts from StyleGAN by \citet{karras2019style} such as a non-linear mapping network and styles that control adaptive instance normalization (AdaIN), we present a novel generative network designed to produce more realistic objects that are robust to pixel-volume outliers and overfitting. This is achieved using 3D deconvolution blocks in the synthesis network, an augmented discriminator based on multiple samples (\citet{lin2018pacgan}), and a Wasserstein loss function with gradient penalty (\citet{arjovsky2017wasserstein}, \citet{GulrajaniAADC17}).


\textbf{Material Microstructures: } The application focus of this architecture is in the field of materials science, which explores the relationships between the processing, structure, and properties of materials.  The interplay between these relationships serve as means for controlling the performance of materials for different applications.  This work focuses on a subset of materials known as crystalline materials.  Crystalline materials are those whose atoms arrange themselves into patterns with long range order.  The most common types of crystalline materials are single crystals and polycrystals.  Single crystal materials are those where a single pattern of long range order is present throughout the entire structure, such as in gemstones, silicon wafers for microchips, and blades found in the hot section of turbine engines.  Polycrystals, which are more prevalent, are materials composed of many connected crystals, called grains, where each grain has local long range order and unique crystal orientation.  The majority of metals and ceramics used in industrial and consumer applications are polycrystalline, including those found in transportation, structural materials, consumer appliances, and aerospace.

In polycrystalline materials, much of the relationship between material processing, structure, and properties manifests through the shape, orientation, and arrangement of the constituent grains.  During solidification or thermal processing, numerous grains form and grow, either through nucleation or diffusive boundary migration, until they impinge upon their grain neighbors, resulting in a fully formed polycrystalline solid.  While the physics of grain formation at thermodynamic equilibrium are relatively well understood, there are many processes that can distort grain shape in a variety of ways through mechanical work, such as rolling and extrusion.  Additionally, some high temperature processing, such as metal joining and additive manufacturing, results in large thermal gradients where thermodynamic equilibrium is not achievable on normal processing timescales.  These extreme processing routes can lead to a variety of different grain shapes and arrangements at the microstructural level, which consequently lead to differences in material response at the property level.  Different grain morphologies can produce changes in stiffness, strength, ductility, and conductivity, all of which can impact material performance in technical applications.  For machine learning to be a viable tool to explore processing-structure-properties relationships in crystalline materials, a critical first step is a generative architecture that can understand grain structure.

In this paper, we make the following contributions:
\begin{enumerate}
    \item We address the 3D object generation problem by proposing 3DMaterialGAN, a novel generative architecture that provides promising results for unsupervised learning of detailed contextual shape information.
    
    \item Our network takes only 3D data as input and outperforms state-of-the art models on the ModelNet benchmark dataset (\citet{Wu2015}) for shape classification through a feature-rich space learned by the discriminator.
    
    \item To show generalizability and application to material science, we demonstrate the use of our network on a large scale 3D dataset of a polycrystalline titanium alloy by \citet{Hemery2019}, which we validate by statistical comparison of moment invariants.
\end{enumerate}

\section{Related Work}


\textbf{3D Shape Completion and Synthesis. }
Until recently, much of the work on 3D object generation included the retrieval of or combining of the components of the 3D object, such as the works of \citet{Chaudhuri2011}, \citet{Funkhouser2004}, \citet{kalogerakis2012probabilistic}. In these cases, given a database of shapes, a probabilistic graphical model will learn the geometric and semantic relationships that will yield a stylistically compatible object. Taking this a step further, \citet{Wu2015} represented geometric 3D shapes as probability distributions of binary variables on a 3D voxel grid and was able to successfully demonstrate shape completion from 2.5D depth maps. A not so similar but related work from \citet{choy20163d} proposed a network that used the ShapeNet dataset to learn a mapping from images to their underlying 3D shapes---given an input image of a plane, generate a 3D representation of a plane. This has since led to works from \citet{Zhang2018Reconstruct, Wu_MarrNetNIPS2017, noguchi2019rgbd} and \citet{nguyen2019hologan} which generate 3D representations from 2D images. Although these showed encouraging results, most of these methods relied on some form of supervision and lack work on the latent space to 3D object generation side. To circumvent this, \citet{sharma2016vconv} proposed a promising autoencoder-based network to learn a deep embedding of object shapes in an unsupervised fashion, which yielded then state-of-the-art shape completion results. 

\textbf{3D GANs for Shapes. } 
The most relevant works are from \citet{Wu_NIPS2016_6096} and \citet{Zhu2018}. The 3D-GAN network from \citet{Wu_NIPS2016_6096} generates 3D objects from a low-dimensional probabilistic latent space, thus enabling the sampling of objects \textit{without} a reference image or CAD models, and the exploration of the 3D object manifold. Similarly, \citet{Zhu2018} propose a novel 3D GAN network, but with the small twist of using a 2D image enhancer. The enhancer network is able to effectively learn and feed the image features into the 3D model generator to synthesize high quality 3D data. While both of these networks deliver on their promise to provide a solution to the 3D model generation problem, they offer limited resolution and detail in the shapes that they generate.  Our proposed network provides a solution to this challenge by generating detailed shapes without the use of additional information from 2D images. To the best of our knowledge, we achieve state-of-the-art performance on the ModelNet benchmark dataset, while also demonstrating the value of our network to solve a real-world problem in materials science. 

\textbf{Generation of Non-Crystalline Material Structures:} 
\citet{Bapst2020} used graph neural networks of 3D atomic arrangements to describe glasses, which have amorphous microstructures.  These graphical descriptions were then used as predictive tools to explore how structure affects mobility and resultant glass properties. Additionally, \citet{Cecen2018} used 3D convolutional neural networks to characterize stochastic microstructures made from filtered noise. Stochastic microstructures have also been investigated in 2D by \citet{Bostanabad2016} using classification trees as well as by \citet{Li2018} using GANs. 

\textbf{Synthetic Crystalline Microstructures: }
Physics-based models have been used for microstructure generation due to their high level of detail and realistic morphologies.  Work by \citet{Coster2005} used a Voronoi tessellation model that simulated ceramic grain boundary evolution based on equations developed by \citet{johnson1939reaction}, \citet{avrami1939kinetics}, and \citet{kolmogorov1937statistical}.  Additionally, the work of \citet{Nosonovsky2009} combined Monte Carlo simulations and grain growth kinetics to model metal crystallization.  These physics-based models show promise, but often require significant computational power as well as detailed knowledge of the energetics of the system.  To avoid this knowledge and computation burden, models have also been developed to generate microstructures based on their statistics.  For example, \citet{quey2011large} as well as \citet{Groeber2014} use statistical descriptors to build microstructures using tessellation and ellipsoid coarsening, respectively.  Additionally, \citet{Callahan2016} compared in detail microstructures generated using the approach of \citet{Groeber2014} with experimental results. Approaches like these are computationally efficient and versatile, compared to physics-based models, but show less realistic grain morphologies and limited accuracy in local grain environment descriptions.

\textbf{Machine Learning with Crystalline Microstructures:  }
 \citet{Liu2015} used structural optimization to explore the microstructure space of the iron-gallium alloy Galfenol.  This approach focused on techniques for optimizing grain arrangement in the Galfenol microstructure to achieve an orientation distribution that would improve desired properties.  While this approach allows for extensive exploration of property space, some challenges arise because the basis of the model is theoretical, so it presents no means of verifying that the proposed microstructures can be physically realized with available processing techniques.  There has, however, been progress with machine learning techniques based on experimental results. Work by \citet{DeCost2017} compiled a database of 2D ultra-high-carbon steel microstructures, and then classified them based on distributions of microstructural features (\citet{DeCost2017a}). Additionally, in work by \citet{Iyer}, a Wasserstein GAN with gradient penalty was used to generate 2D microstructures using the database of \citet{DeCost2017} as training data.  More closely related to this work, \citet{Fokina2020} used StyleGAN to generate various 2D microstructures, and \citet{Cang2017} used a convolutional deep belief network to generate 2D microstructures of the same titanium alloy investigated in this study.  While approaches like this have good experimental foundation, validation of microstructures is difficult in these 2D cases.  The challenge in 2D microstructure assessment arises from the fact that unlike conventional photographs, material micrographs are planar images of solid material, and offer little perspective-based information from which 3D appearance can be inferred.  More explicitly, \citet{torquato2002random} show that for certain shape classes such as anisotropic materials, establishing 3D structure from 2D is mathematically intractable.  In order for output from machine learning-based models to be fully comparable to material microstructures, generation and evaluation needs to be based in 3D space. 

\section{Model}

In this section, we introduce the foundational idea of generative adversarial networks inspired by the StyleGAN network, followed by a formal outline of our proposed 3DMaterialGAN network.

\textbf{Generative Adversarial Network (GAN):  }  
First proposed by \citet{Goodfellow_NIPS2014_5423}, a simple generative adversarial network consists of a generator $G$ and a discriminator $D$. The generator tries to synthesize samples that look like the training data, while the discriminator tries to determine whether a given sample is a real sample originated from the ground truth data or from the generator. The discriminator $D$ outputs a confidence value $D(x)$ of whether input $x$ is real or synthetic.

\textbf{StyleGAN Network Architecture:  } 
StyleGAN, developed by \citet{karras2019style}, garnered widespread attention for its life-like image quality and unsupervised high-level attribute separation in the generated output. Because of these characteristics, we used StyleGAN as the base architecture for our network. Instead of passing a random noise vector $z$ to the generator, $z$ is first mapped to an intermediate latent space $W$, which is transformed into spatially invariant styles $\mathbf{y} = (\mathbf{y}_s, \mathbf{y}_b)$. This is then used to control the generator through adaptive instance normalization (AdaIN) at each convolutional layer. The AdaIN is defined as: 

 \begin{align}
     \text{AdaIN}(\mathbf{x}_i, \mathbf{y}) = \mathbf{y}_{s,i} \frac{\mathbf{x}_i - \mu(\mathbf{x}_i)}{\sigma(\mathbf{x}_i)} + \mathbf{y}_{b, i} 
\end{align}
 Here, $\mathbf{x}_i$  is a feature map normalized separately, and then scaled and biased using the corresponding scalar components from style $\mathbf{y}$.
 
\begin{figure}[t]
   \centering
\includegraphics[width=0.95\textwidth]{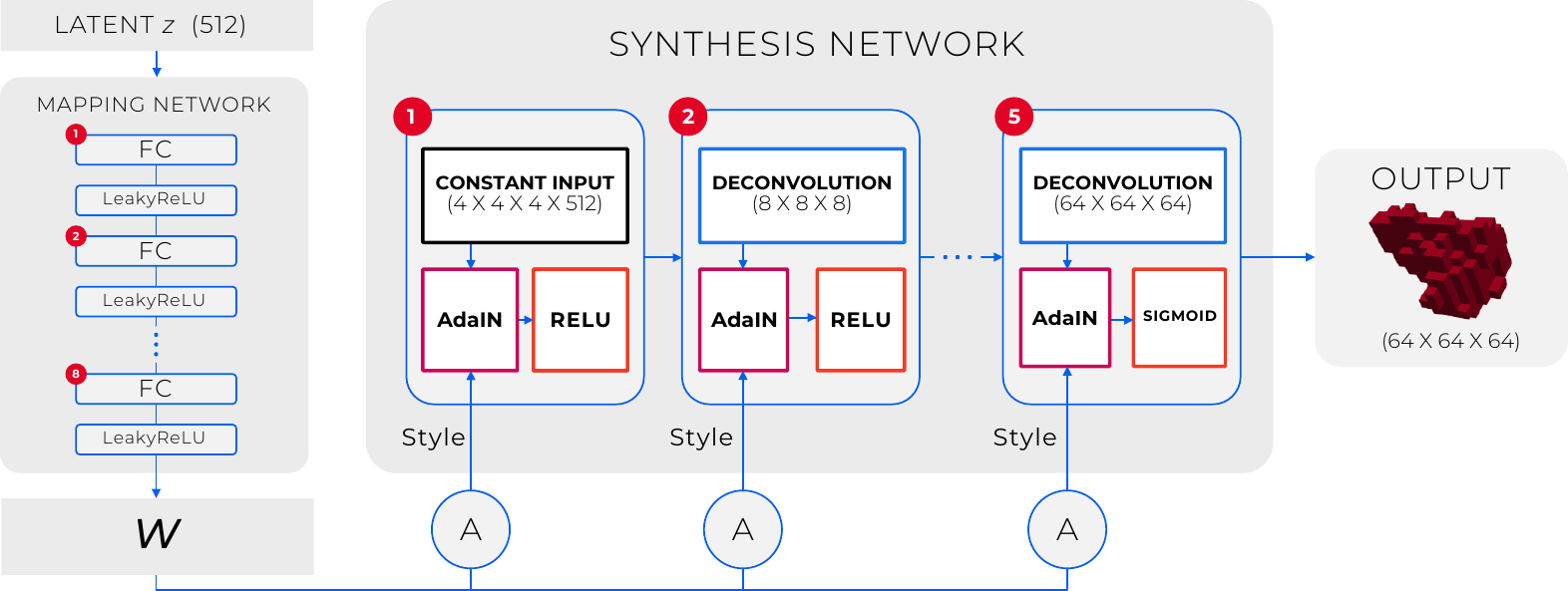}
\caption{\textsc{Generator Architecture of 3DMaterialGAN.} A mapping network comprised of eight Fully Connected layers (FC) with Leaky ReLU activation function after each FC layer takes as input a 512 dimensional latent vector $z$. The output is then mapped to an intermediate latent space $\mathcal{W}$, converted into styles using a learned affine transformation (A), and passed through an AdaIN operation for each of the five blocks in the synthesis network. Block 1 passes constant input through AdaIN and ReLU activation functions, while Blocks 2-5 are deconvolution blocks progressively grown from $8 \times 8 \times 8 \to 64 \times 64 \times 64$. }
\label{fig:arch}
\end{figure}

\textbf{3DMaterialGAN Network Architecture: }
In our 3D Material Generative Adversarial Network (3DMaterialGAN), an initial 512-dimensional latent vector $z$ is chosen through probabilistic sampling of latent space.  The generator $G$ then maps this latent vector into an intermediate latent space $\mathcal{W}$ using a mapping network of 8 fully connected layers with a LeakyReLU activation function after each layer. The output of $\mathcal{W}$ is then converted into styles using learned affine transformation, denoted as $A$ in Figure \ref{fig:arch}, and these styles are used to control adaptive instance normalization (AdaIN) in the synthesis network. This localization of styles better preserves the fine-scale information of each object, allowing for more detailed shape capture compared to other networks (\citet{Wu_NIPS2016_6096, Zhu2018}). Our synthesis architecture diverges from that of StyleGAN by using blocks consisting of a 3D deconvolution passed through an AdaIN operation and ReLU activation function. The synthesis network has a total of five blocks, with the first having a constant input vector and normalization, and the fifth having a sigmoid activation function instead of a ReLU. In comparison, StyleGAN blocks use upsampling followed by two alternating convolution layers and AdaIN operations, with noise introduced after each convolution to add stochastic variation. However, in 3D, this noise was found to add instability during training, and was therefore not included in the 3DMaterialGAN network. The discriminator uses five 3D convolution layers, each with a LeakyReLU activation functions except for the last layer. The discriminator makes decisions based on multiple samples from the same class, either real or generated, as discussed by \citet{lin2018pacgan}.

\textbf{ShapeNet Dataset:  }
From the ShapeNet Dataset by \citet{chang2015shapenet}, six major object categories were used in the network training (car, chair, plane, guitar, sofa, rifle). Each category has a 128-sample training set. Objects are in a voxel-based format which has dimension of $64 \times 64 \times 64$. 

\textbf{Titanium Dataset:} The material data was gathered by J. Wendorf and A. Polonsky (\citet{Hemery2019}) using a method  developed by \citet{Echlin2015} known as the Tribeam system, which performs rapid serial sectioning using ablation with ultrashort pulse femtosecond lasers.  The material of interest is a polycrystalline wrought titanium alloy containing 6.75 wt.\% aluminum and 4.5 wt.\% vanadium.  This alloy, commonly referred to as titanium 6-aluminum 4-vanadium, or Ti 6-Al 4-V, has applications in turbine engines, aerospace, and medical prostheses.  The grain information is gathered at the voxel level using electron backscatter diffraction, with each voxel having a size of 1.5 $\mu$m $\times$ 1.5 $\mu$m $\times$ 1.5 $\mu$m. A total of 84,215 grains from the dataset were used for network training. Each grain is passed to the network inside a cubic volume of size $64 \times 64 \times 64$ voxels.

\textbf{Training details: }
Stable training was achieved by setting the learning rate for both the generator and discriminator at 0.0002, with a batch size of 16. The discriminator is updated five times for each generator update and employs the Adam optimizer described by \citet{kingma2014adam} with $\beta=0.5.$ The output of the network is in a 64 $\times$ 64 $\times$ 64 dimensional space. The Wasserstein loss as described in \citet{arjovsky2017wasserstein} with a gradient penalty is used for the discriminator and the generator, which are defined as follows:

\begin{align}
L_{\text{D-Loss}} &= - \frac{1}{m} \sum_{i=1}^{m} D_\phi \left( x^{(i)} \right) + \frac{1}{m} \sum_{i=1}^{m} D_\phi \left( G_{\theta}\left(z^{(i)} \right) \right) +\frac{1}{m} \sum_{i=1}^{m} \lambda \left( \left\lVert\nabla_{\hat{x}^{(i)} }D_\phi\left( \hat{x}^{(i)} \right)\right\rVert_2  - 1 \right)^2 \\
L_{\text{G-Loss}} &= - \frac{1}{m} \sum_{i=1}^{m} D_{\phi} \left( G_{\theta}\left(z^{(i)}\right)  \right)   
\end{align}
Here, $\hat{x}^{(i)}$ is defined as \(\epsilon x^{(i)} + (1 - \epsilon)G_{\theta}(z^{(i)}) \) and $\lambda$ is a gradient penalty coefficient. \(\epsilon \) is a uniform random variable in [0,1]. 
\({x^{(i)}}_{i=1}^{m} \sim \mathbb{P}_r \) is a batch from real data, 
\({z^{(i)}}_{i=1}^{m} \sim p(z) \)  is a batch of random noise, and $m$ denotes batch size. $G_{\theta}$ is a generator network and $D_{\phi}$ is a discriminator network.

\section{Evaluation}

We evaluate our network across several areas. First, we show qualitative results of generated 3D objects from the ShapeNet dataset by \citet{chang2015shapenet}. Then, we evaluate the unsupervised learned representations from the discriminator by using them as features for 3D object classification. Lastly, quantitative results are shown on the popular benchmark ModelNet dataset from \citet{Wu2015}.

\textbf{3D Object Generation:}
We train one 3DMaterialGAN for each object category using a 512 dimensional random vector which follows a normal distribution with mean 0 and variance 0.2. We compare our generated objects with \citet{Wu_NIPS2016_6096}, because \citet{Zhu2018} used an  enhancer network and additional information from 2D rendered images during training. Our network synthesizes high-resolution 3D objects ($64 \times 64 \times 64$) with detailed shape information trained from only 3D input. And to ensure that our network is not memorizing the training data, we perform nearest neighbor analysis for the synthesized objects as in \citet{Wu_NIPS2016_6096}. Features were calculated from both the generated samples and training samples to find the nearest neighbour using $\ell_2$ distance. This analysis yielded that the generated samples were not identical to its nearest neighbor.

\textbf{3D Object Classification:}
To evaluate the unsupervised learned features from our network, we use the method in \citet{Wu_NIPS2016_6096} to provide a means of comparison, as there is no established standard.  We train our 3DMaterialGAN network on six object categories (bed, car, chair, plane, sofa, table), one object (gun) less than the analysis done by \citet{Wu_NIPS2016_6096}. Each object category has 25 samples, same as \citet{Wu_NIPS2016_6096}, in the training set from the ShapeNet dataset. 

Next, we use the ModelNet dataset from \citet{chang2015shapenet} to evaluate the unsupervised features learned by our network. The ModelNet dataset has two categories: ModelNet10 (10 classes) and ModelNet40 (40 classes). ModelNet10 has a total of 3991 training samples and 908 test samples. From this dataset, we used 100 samples from each category, totaling 1000 training samples. ModelNet40 has a total of 9843 training samples and 2468 test samples. From this dataset, we used 100 samples from each category when available, as some categories had less than 100 samples, totaling 3906 training samples. For ModelNet10 and ModelNet40, we used the entire test dataset provided, and \textit{less training samples} than what \citet{Wu_NIPS2016_6096} and \citet{Zhu2018} used for their evaluation.

To provide a fair comparison, we used the same kernel size = \{8, 4, 2\} defined by both \citet{Wu_NIPS2016_6096} and \citet{Zhu2018}, as well as the defined stride = \{4, 2, 1\}  by \citet{Zhu2018}. We calculate features from the second, third and fourth layers of the trained discriminator, which are then concatenated after applying max-pooling with the defined kernel size and stride. We then train a linear Support Vector Machine (SVM) on training features and calculate classification accuracy on the test features.

\textbf{Evaluation of Generated Grains:}
Grain formation in polycrystalline microstructures is influenced by a combination of processing, thermodynamics, and statistics.  Because of this, evaluation of grains cannot be easily done by visual evaluation or direct object comparison. For this reason, we use statistical distribution comparison of 3D moment invariants to assess the quality of generated results for single grains.  Much like the image moments used in 2D analysis, 3D moment invariants are integration-based descriptors that numerically quantify an object based on the distribution of its solid volume.  These types of invariants have been used previously by \citet{Kazhdan2003} as shape descriptors for general 3D objects and in materials science by  \citet{MacSleyne2008} and \citet{Trenkle2020} to describe the shapes of particles such as grains and inclusions.  Following the approach of \citet{MacSleyne2008}, Cartesian moment descriptors $\mu_{pqr}$ take the form:
\begin{align}
    \mu_{pqr} = \int\int\int x^p y^q z^r F(r) dr
\end{align}
where $F(r)$ is a characteristic function that has a value of one in material regions and a value of zero in void regions.  In this study, because all ground-truth data is in voxel-based form, all integrations are done as Riemann sums in voxel space, rather than as continuous integrals. Following this Cartesian moment form, $\mu_{000}$ is directly equal to the volume $V$ of an object, and the centroid $(X_c, Y_c, Z_c)$ of an object can be expressed in the following form:
\begin{align}
    (X_c, Y_c, Z_c) = \left(\frac{\mu_{100}}{V}, \frac{\mu_{010}}{V}, \frac{\mu_{001}}{V} \right)
\end{align}
Working from a coordinate space originating at the centroid, the non-normalized moment invariants $\mathcal{O}_1, \mathcal{O}_2, \mathcal{O}_3$ are as follows:
\begin{align}
    \mathcal{O}_1 &= \mu_{200} + \mu_{020} + \mu_{002} \\
    \mathcal{O}_2 &= \mu_{200}\mu_{020} + \mu_{200}\mu_{002} + \mu_{020}\mu_{002} - \mu_{110}^2 - \mu_{101}^2 - \mu_{011}^2 \\
    \mathcal{O}_3 &= \mu_{200}\mu_{020}\mu_{002} + 2\mu_{110}\mu_{101}\mu_{011} - \mu_{200}\mu_{011}^2 - \mu_{020}\mu_{101}^2 - - \mu_{002}\mu_{110}^2
\end{align}
These can then be normalized to object volume to produce the moment invariants $(\Omega_1, \Omega_2, \Omega_3)$ considered in this study.
\begin{align}
   \left( \Omega_1, \Omega_2, \Omega_3 \right) = \left( \frac{3V^{5/3}}{\mathcal{O}_1}, \frac{3V^{10/3}}{\mathcal{O}_2}, \frac{V^{5}}{\mathcal{O}_3} \right) 
\end{align}
The distributions of these moment invariants are used to evaluate the shape quality of the generated grains.  Additional analysis of 3D moment invariants can be found in \citet{MacSleyne2008}.

\section{Results}

\begin{figure}[t]
   \centering
    \includegraphics[width=.9\textwidth]{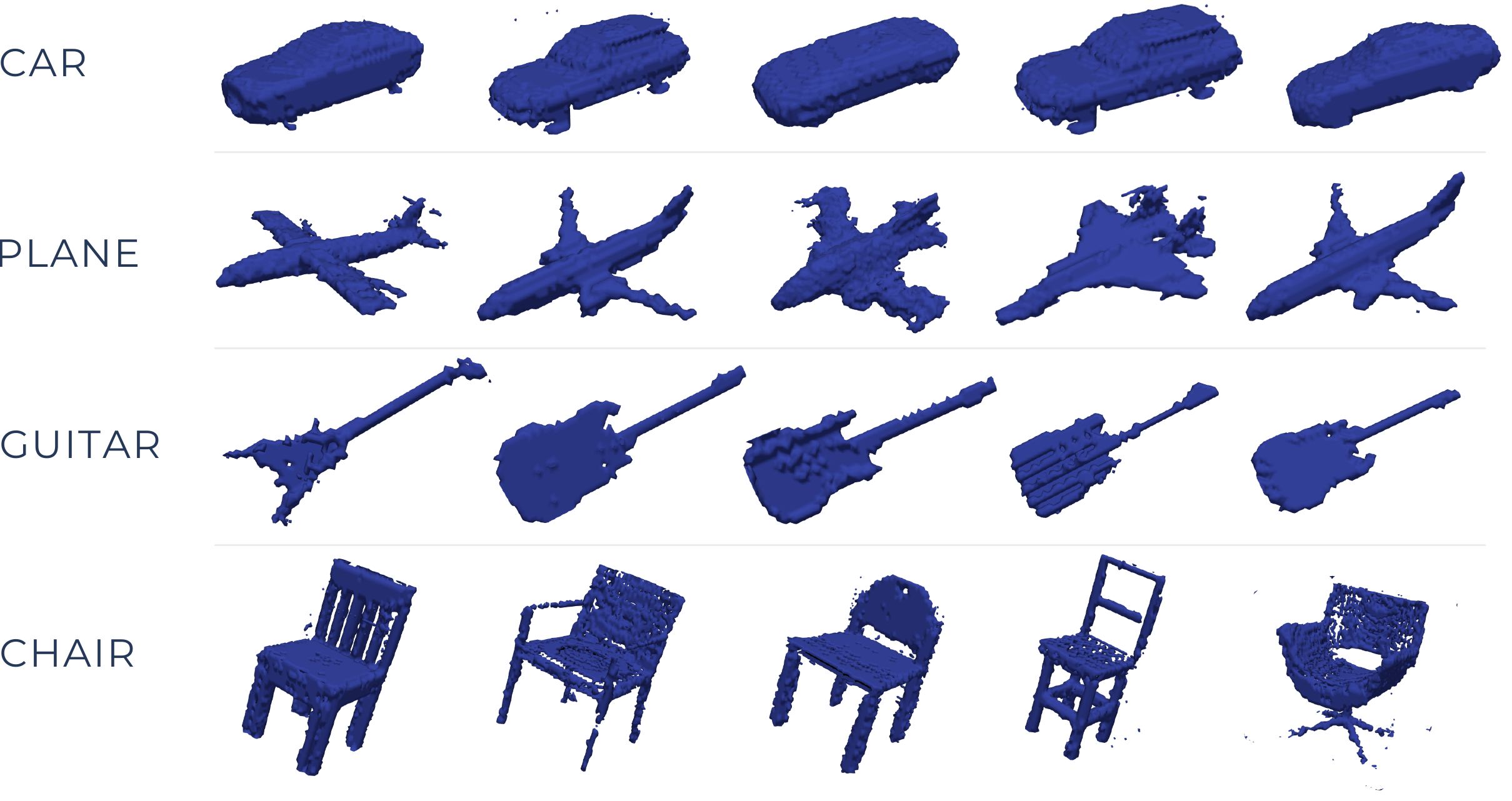}
    \caption{\textsc{Shapenet Output. } Here we show a diverse range of 3D objects with detailed shape information generated by our 3DMaterialGAN network.}
    \label{fig:shapenet_output}
\end{figure}
\textbf{ModelNet Benchmark Results:} With fewer training samples than \citet{Wu_NIPS2016_6096} and \citet{Zhu2018}, we achieve 92.29\% accuracy on the ModelNet10 dataset and 85.08\% accuracy on the ModelNet40 dataset. Given the graph in \textbf{Figure 4 and Figure A1 } in the supplementary material from \citet{Wu_NIPS2016_6096}, if we use a comparable training set size, we achieve a \textbf{2.29\%} improvement on ModelNet10 and a \textbf{3.78\%} improvement on ModelNet40 dataset. Furthermore, \citet{Zhu2018} uses all available training samples as well as additional rendered 2D images for both datasets, compared to our use of both fewer training samples and only 3D input. 

\begin{tabular}{l|c|r|r}
      \toprule 
      \textbf{Method} & \textbf{Supervised}   & \textbf{ModelNet10} (\%) & \textbf{ModelNet40} (\%) \\
      \midrule
      3D ShapeNets(\citet{Wu2015}) & Yes & 83.54 & 77.32 \\
      VoxNet (\citet{maturana2015}) & Yes & 92.00 & 83.00\\
      Geometry Image (\citet{Sinha2016DeepL3})    & Yes & 88.40 & 83.90\\
      PointNet (\citet{qi2017pointnet}) & Yes & 77.60 & - \\
      GIFT (\citet{bai2016gift}) & Yes & 92.35 & 83.10 \\
      FusionNet (\citet{hegde2016fusionnet}) & Yes & 93.11 & 90.80 \\
      \midrule
      SPH (\citet{sph2003}) & No & 79.79 & 68.23 \\
      LFD (\citet{chen2003visual}) & No & 79.87 & 75.47 \\
      VConv-DAE (\citet{sharma2016vconv}) & No & 80.50 & 75.50 \\
      3D-GAN (\citet{Wu_NIPS2016_6096}) & No & 91.00 & 83.30 \\
      3D-GAN (\cite{Wu_NIPS2016_6096}) ($\approx$ \textbf{100} samples) & No & 90.00 & 81.30 \\
      \textbf{Ours (3DMaterialGAN)} (\textbf{100} samples)  & \textbf{No} & \textbf{92.29} & \textbf{85.08} \\ 
      
      \bottomrule
\end{tabular}\\

\begin{figure}[t]
   \centering
    \includegraphics[width=0.9\textwidth]{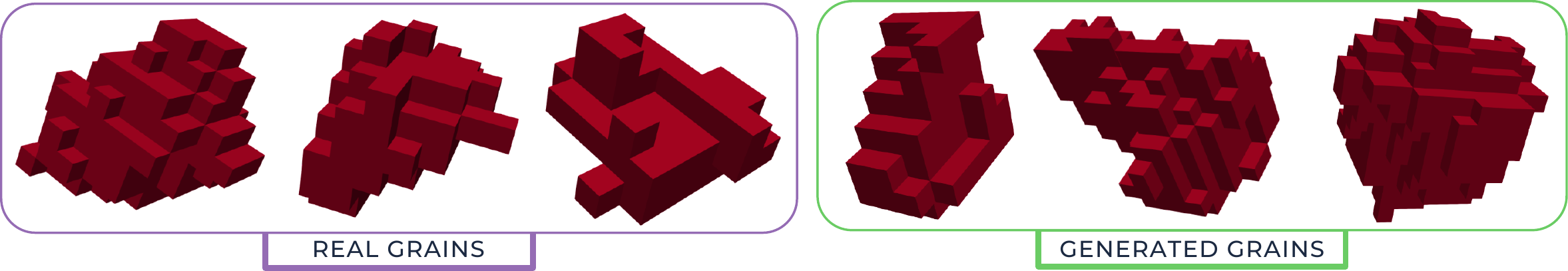}
    \caption{\textsc{Example of Single Grains. } Voxel size is 1.5 $\mu$m $\times$ 1.5 $\mu$m $\times$ 1.5 $\mu$m.  For generated grains, only largest connected component is shown.}
    \label{fig:single_grains}
\end{figure}

\textbf{Grain Generation Results: }  Representative volumes of grains generated using the 3DMaterialGAN are shown in Figure \ref{fig:single_grains}.  Unlike the benchmark results, which are shown as contour displays, these figures are shown in voxel form, which matches the representation of the ground-truth data.  It should be noted that some generation artifacts were present in 64 $\times$ 64 $\times$ 64 volume containing these grains, with the most common of these artifacts being voxels of solid volume at random locations in the void.  Some examples of these artifacts can be seen in the contour images of the benchmark set shown in Figure \ref{fig:shapenet_output}.  In the case of grain volumes shown in Figure \ref{fig:single_grains}, only the largest connected volume is shown.  These representative volumes show  shapes that would be reasonably expected of grains in an equiaxed polycrystalline metal like wrought Ti 6-Al 4-V, but these representative volumes alone are insufficient verification of generated grain quality.

\begin{figure}[t]
    \centering
    \includegraphics[width=0.95\textwidth]{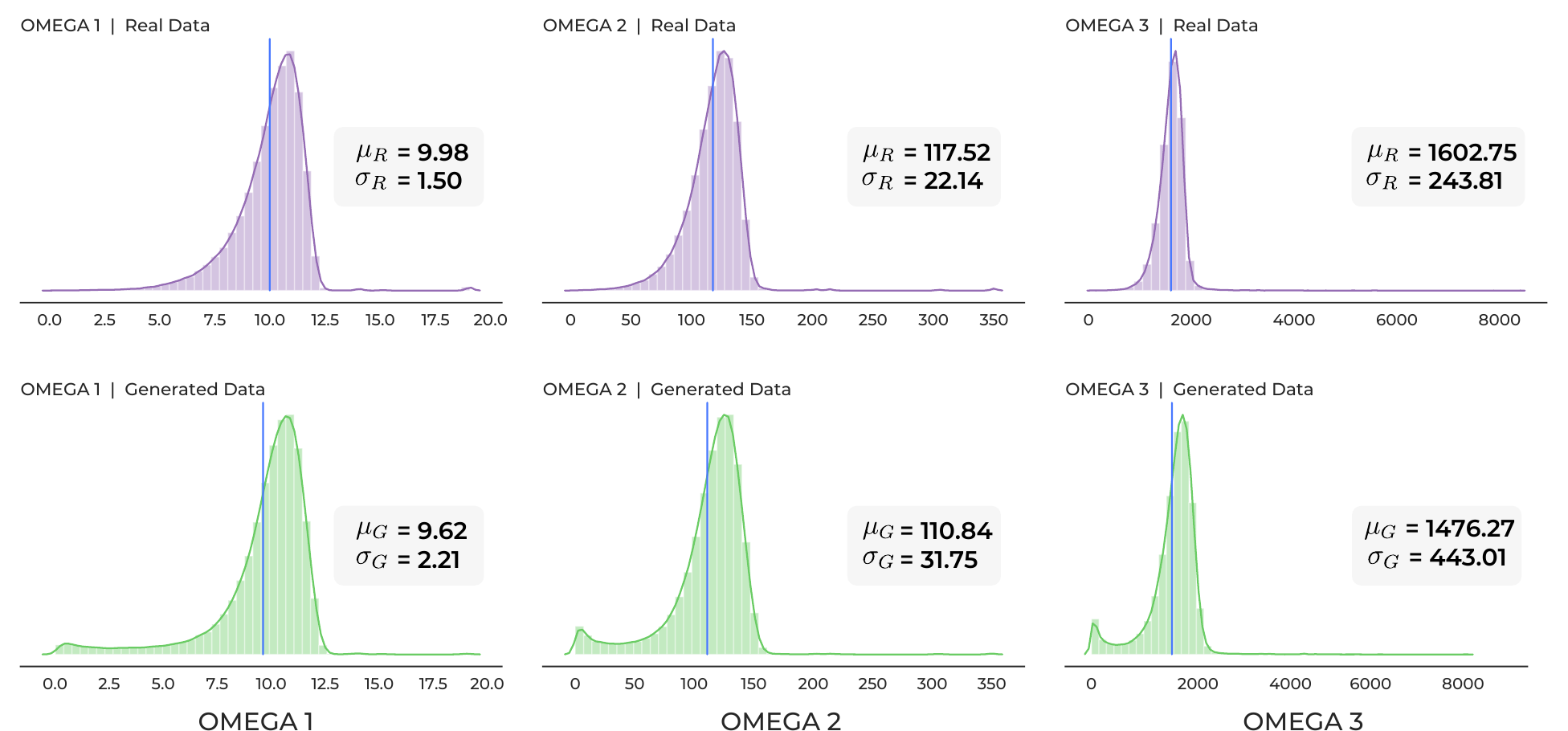}
    \caption{Histogram of moment invariants($\Omega_1, \Omega_2, \Omega_3$) for grains. Vertical lines indicate the mean.}
    \label{fig:moment_invariants}
\end{figure}

The distributions of the 3D moment invariants $(\Omega_1, \Omega_2, \Omega_3)$ evaluated in this study are shown in Figure \ref{fig:moment_invariants}.  The ground truth dataset contained 84,215 grains and the 3DMaterialGAN generated a total of 150,000 grains.  For each of the three invariants, less than 0.2\% of the grains had values that were either infinite or nonphysical across both the ground truth and generated sets. In observed cases, this resulted from very small grain structures that were linear or planar in nature, which led to extremely large errors in Riemann summation during calculation of the invariants.  Due to their nonphysicality, these values were omitted from the statistical distribution comparisons.

The ground-truth mean ($\mu_R$) and standard deviation ($\sigma_R$) are compared with the generated mean ($\mu_G$) and standard deviation ($\sigma_G$) in the insets of Figure \ref{fig:moment_invariants}.  These values show a trend of differences between the real and generated data, with the generated data consistently exhibiting lower mean values and larger standard deviations.  Comparison of the distribution shapes in Figure \ref{fig:moment_invariants} reveals the cause to be that the generated data has a range of moment invariants values in the region close to zero not seen in the real dataset. These values are likely due to the previously mentioned generator noise.  Although the generator noise was not displayed in Figure \ref{fig:single_grains} for visual clarity, this noise was still included in the moment invariant calculation, which was performed on raw network output to minimize bias in quantitative comparison.  This distinction is critical because, as Figure \ref{fig:single_grains} shows, some of the grains are relatively small in volume compared to the 64 $\times$ 64 $\times$ 64 voxel region in which they were generated.  Thus, generator artifacts located far away from the grain itself can cause significant shifts in the centriod location that is used to establish the coordinate basis for the calculation of the moment invariants, creating distortions in the data.  Beyond this effect, the overall shape and position of the main peaks for both distributions are very similar, indicating that the 3DMaterialGAN network is producing data that is similar to the ground truth data without directly replicating it. 

\section{Conclusion}

\textbf{Summary: } This work presents 3DMaterialGAN, a network designed for use on crystalline material microstructures that can recognize and synthesize high-resolution 3D objects from latent space vectors without any supplemental 2D input. We demonstrate the capabilities of this network on the ModelNet benchmark dataset, as well as on an experimental 3D material dataset of Ti 6-Al 4-V.  On the benchmark dataset, the 3DMaterialGAN yields more reliable discriminator classification with exposure to fewer objects, which highlights the quality of discriminator feature recognition in the network.  When working with experimental data, the effects of generator noise produce some variations in moment invariant distribution, but the 3DMaterialGAN network is still able to produce results morphologically similar to the ground truth without being a direct replication.

\textbf{Broader Impact: } In the context of impact on the scientific world at large, this work has several implications centered in materials research and development. Since the 1990s, there have been several large-scale initiatives, including the Materials Genome Initiative (MGI) developed in the United States under the Obama Administration (\citet{national2011materials}), with focus to improve the efficiency of the materials development process.  Therefore, many applications across numerous industries rely on metal alloys and ceramic compositions that are over 50 or even 100 years old.  Development of new materials that are suitable for a wide range of processing approaches and applications is an effort that costs an enormous amount of time, capital, and manpower.  A major contributor to this cost is the enormous price of iteration: any new material composition must be evaluated under all performance scenarios of interest, and this must be done enough times to ensure the evaluation results are representative of all potential processing variabilities.  Turbine materials must withstand the environment of an engine, and biomedical implants must suitable for a lifetime of use inside the body. Evaluations of these materials must be robust enough to be representative for thousands or even millions of manufacturing iterations.  The magnitude of this development cost means that any progress made towards reducing it presents the potential for significant technical advances in applications like power generation, transportation, construction, consumer products, aerospace, and defense.  

This work is the first step in teaching a neural network to understand the crystalline microstructure of a titanium alloy in full detail, backed by experimental verification.  If a network can understand the morphology of titanium grains and how those grains are arranged with respect to one another for different processing conditions, it has the potential to generate new titanium microstructures for evaluation that are representative of a given processing condition, which reduces the cost of iteration needed for building a representative description of the titanium alloy.  With enough information, the network may be able to predict how changes in processing conditions affect the titanium microstructure without performing any new experimental evaluations.  Even this singular problem is a major challenge, but this work helps establish an approach through which this challenge might be overcome.

\section{Acknowledgements}
This research is supported in part by NSF awards number 1934641 and 1664172. The authors gratefully acknowledge Andrew Polonsky and Joseph Wendorf for collection of the 3D Ti 6-AL-4-V datasets. The MRL Shared Experimental Facilities are supported by the MRSEC Program of the NSF under Award No. DMR 1720256; a member of the NSF-funded Materials Research Facilities Network (www.mrfn.org).

\small
\bibliography{sources}

\end{document}